\pdfoutput=1

\documentclass{article}
\usepackage{threeparttable}

\usepackage[final]{acl}
\usepackage{acl}
\usepackage{array}
\usepackage{tabularx}
\usepackage{cleveref}
\usepackage{booktabs}
\usepackage{graphicx}

\usepackage{times}
\usepackage{latexsym}

\usepackage[T1]{fontenc}

\usepackage[utf8]{inputenc}

\usepackage{microtype}

\usepackage{inconsolata}

\usepackage{graphicx}

\title{Parallel Corpora for Machine Translation in Low-Resource Indic Languages: A Comprehensive Review}

\author{Rahul Raja \\
  Carnegie Mellon University \\
  Stanford University\\
  LinkedIn\thanks{Work does not relate to position at LinkedIn.} \\
  \\\And
  Arpita Vats \\
  Boston University \\
  Santa Clara University \\
  LinkedIn\footnotemark[1] \\
  }

\begin{document}
\maketitle
\begin{abstract}
Parallel corpora play an important role in training machine translation (MT) models, particularly for low-resource languages where high-quality bilingual data is scarce. This review provides a comprehensive overview of available parallel corpora for Indic languages, which span diverse linguistic families, scripts, and regional variations. We categorize these corpora into text-to-text, code-switched, and various categories of multimodal datasets, highlighting their significance in the development of robust multilingual MT systems. Beyond resource enumeration, we critically examine the challenges faced in corpus creation, including linguistic diversity, script variation, data scarcity, and the prevalence of informal textual content. We also discuss and evaluate these corpora in various terms such as alignment quality and domain representativeness. Furthermore, we address open challenges such as data imbalance across Indic languages, the trade-off between quality and quantity, and the impact of noisy, informal, and dialectal data on MT performance. Finally, we outline future directions, including leveraging cross-lingual transfer learning, expanding multilingual datasets, and integrating multimodal resources to enhance translation quality. To the best of our knowledge, this paper presents the first comprehensive review of parallel corpora specifically tailored for low-resource Indic languages in the context of machine translation.
\end{abstract}

\section{Introduction}
\label{sec:intro}
\subsection{Importance of parallel corpora}
Parallel corpora are collections of texts that contain sentence-aligned translations across two or more languages ~\cite{brown}. These resources play a fundamental role in machine translation (MT), cross-lingual natural language processing (NLP), and linguistic research. Unlike monolingual corpora, parallel corpora enable direct learning of translation mappings, making them essential for training statistical and neural MT models ~\cite{koehn2020parallel}.\\
Parallel corpora have been crucial in the development of MT models, starting from phrase-based statistical MT (SMT) ~\cite{romdhane-etal-2014-phrase} to modern neural MT (NMT) approaches ~\cite{stahlberg2020neuralmachinetranslationreview}. In SMT systems, they provided the necessary data for learning phrase alignments and translation probabilities ~\cite{voita-etal-2021-language}. With the rise of transformer-based NMT models ~\cite{vaswani2023attentionneed}, large-scale parallel corpora have become even more critical, as these models rely on extensive aligned data to learn high-quality translation representations.\\
Beyond MT, parallel corpora are used in cross-lingual NLP tasks such as multilingual word embeddings ~\cite{conneau2020unsupervisedcrosslingualrepresentationlearning}, zero-shot learning ~\cite{Artetxe_2019}, and multilingual question-answering systems ~\cite{hu2024gentranslatelargelanguagemodels}. These resources allow models to generalize across languages by leveraging shared semantic representations learned from translation pairs.\\
For low-resource languages, parallel corpora are not just tools for MT but also serve a crucial role in language preservation and revitalization ~\cite{hu-etal-2024-learning}. Many Indic languages lack digitized linguistic resources, making them vulnerable to digital extinction. Creating high-quality parallel datasets ensures that these languages remain computationally accessible, enabling future educational tools, digital assistants, and automated translations ~\cite{anastasopoulos-etal-2020-tico}.
\begin{figure}[h!]
\begin{minipage}[b]{1.0\linewidth}
  \centering
  \centerline{\includegraphics[width=7.7cm]{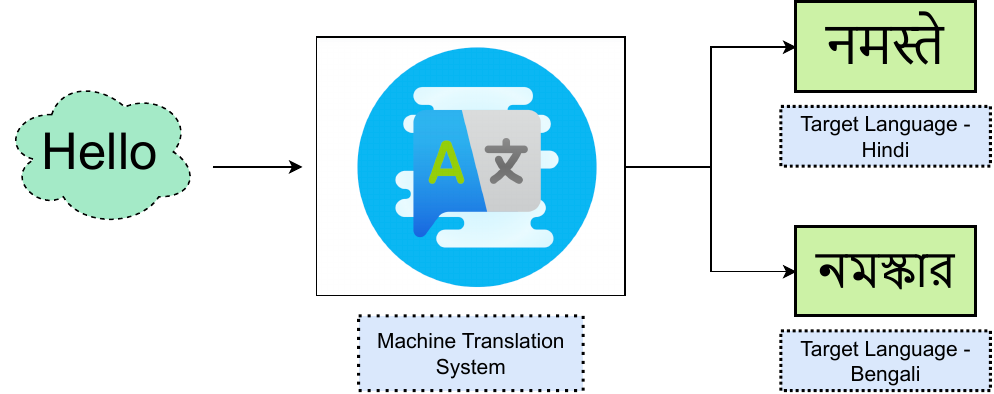}}
  \caption{\label{fig_overview} Overview of Machine Translation Model.}
\end{minipage}
\end{figure}
\subsection{Machine Translation for Indic languages}
\begin{figure*}[htb]
\begin{minipage}[b]{1.0\linewidth}
  \centering
  \centerline{\includegraphics[width=16.5cm]{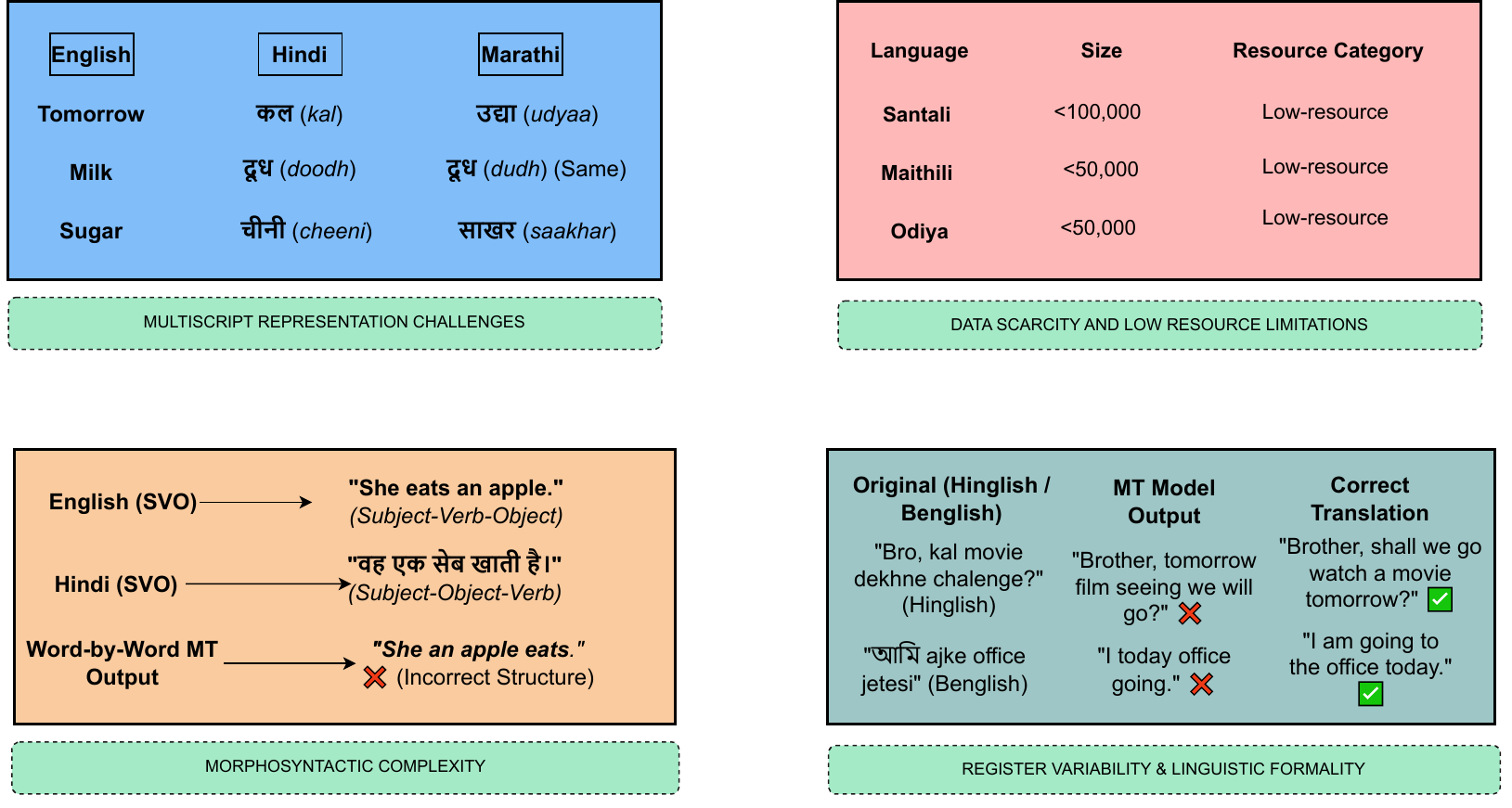}}
  \caption{\label{fig_framework} Challenges in Indic Machine Translation: Key issues include morphosyntactic complexity, script variations, low-resource languages, and translation errors in Hinglish and Benglish.}
\end{minipage}
\end{figure*}

MT for Indic languages faces numerous challenges due to their linguistic diversity, script variations, and resource constraints ~\cite{bala}. Unlike high-resource languages, many Indic languages suffer from limited parallel corpora, making it difficult to train robust translation models. The diversity in syntax and phonology across language families further complicates alignment and translation tasks. The presence of multiple scripts and a lack of standardized transliteration mechanisms hinder effective corpus development. ~\Cref{fig_framework} shows the overview of challenges in Indic MT
\subsubsection{Morphosyntactic Complexity and Linguistic Variability}
Indic languages exhibit significant linguistic diversity, primarily categorized into Indo-Aryan and Dravidian language families ~\cite{masica1993indo}. Indo-Aryan languages, such as Hindi, Bengali, and Marathi, are characterized by inflectional morphology and a relatively flexible subject-object-verb (SOV) ~\cite{SCHOUWSTRA2014431} word order, whereas Dravidian languages, including Tamil, Telugu, and Kannada, employ agglutinative morphology, where words are formed by adding multiple affixes to a root word. These structural differences pose challenges in MT systems, as segmentation strategies that work for one language family may not be effective for another ~\cite{rama2012morphological}. Additionally, phonological distinctions, such as retroflex consonants in Dravidian languages that are absent in many Indo-Aryan languages, complicate speech-to-text and transliteration tasks  ~\cite{annamalai2006language}.

\subsubsection{Multiscript Representation and Orthographic Challenges }
\label{subsec:challenges}
Indic languages are written in multiple scripts, which significantly impact corpus creation and text normalization ~\cite{manohar2024lostnormalizationexploringpitfalls}, ~\cite{hellwig2010interaction}. For instance, Hindi and Marathi share the Devanagari script, but differences in spelling conventions and phonetic representations require preprocessing before effective alignment ~\cite{hellwig2010interaction}. Bengali and Assamese use the Bengali-Assamese script, while Tamil, Telugu, and Kannada have distinct scripts with unique grapheme-to-phoneme mappings ~\cite{gales2007application}.We have also created an indic language categorization illustrated in ~\Cref{Tbl:indic-scripts}. The lack of script standardization introduces inconsistencies in parallel corpora, making text alignment a challenging task. Moreover, the development of optical character recognition (OCR) tools for Indic scripts remains an ongoing challenge, as many scripts have complex ligatures and diacritic variations that reduce OCR accuracy, further limiting the availability of digitized resources for MT ~\cite{sengupta2019ocr}.These script-specific challenges underscore the need for robust preprocessing pipelines and script-aware normalization techniques to improve the quality and usability of Indic language corpora.

\subsubsection{Data Scarcity and Low-Resource Limitations}
The scarcity of high-quality parallel corpora remains a significant obstacle in developing robust MT models for Indic languages ~\cite{10.1145/3587932}. While languages like Hindi and Bengali have relatively larger corpora, low-resource languages such as Santali, Maithili, and Konkani lack sufficient parallel data, restricting the effectiveness of data-driven MT approaches. The limited availability of bilingual datasets hampers the training of neural MT models, which require vast amounts of parallel text for effective generalization. To mitigate this issue, researchers have explored synthetic data generation techniques such as back-translation and cross-lingual transfer learning. However, these approaches often introduce artifacts that can degrade translation quality, highlighting the need for well-annotated, human-verified corpora to support low-resource Indic language MT ~\cite{sengupta2019ocr}.
\subsubsection{Register Variability and Linguistic Formality}
Most existing parallel corpora for Indic languages are derived from formal sources such as news articles, religious texts, and government documents, which do not capture the informal and conversational aspects of language used in everyday communication ~\cite{post2012newsmt}. This imbalance affects the performance of MT systems in real-world applications, as they struggle to translate colloquial expressions, dialectal variations, and code-switched text commonly found in social media and user-generated content ~\cite{rijhwani2020handling}. Code-mixing ~\cite{khanuja2020gluecos}, particularly in Hindi-English and Bengali-English, presents additional challenges, as standard MT models are not optimized for handling intra-sentential language switching ~\cite{pratapa2018code}. The need for diverse corpora that encompass both formal and informal registers is essential to improve translation accuracy across different linguistic contexts.

\section{Parallel Corpora for Indic Languages: Modalities and Comparisons}
The development of MT systems for Indic languages heavily relies on the availability of high-quality parallel corpora. These corpora serve as the foundation for training neural MT models, aligning linguistic structures across languages, and enabling multilingual applications such as automatic translation, speech recognition, and multimodal understanding. Given the diverse nature of Indic languages and their applications in different contexts, parallel corpora can be classified into various types based on the modality of data they contain. This section provides an overview of different types of parallel corpora available for Indic languages, their characteristics, and their significance in MT research.
\subsection{Text-to-Text Parallel Corpora}
Text-to-text parallel corpora form the backbone of MT systems, consisting of bilingual or multilingual sentence-aligned datasets that provide direct translations between languages ~\cite{degibert2024newmassivemultilingualdataset}. These corpora are essential for training supervised MT models and are widely used in both statistical and neural machine translation (NMT) frameworks ~\cite{raunak2024instructionfinetuningneuralmachinetranslation}. 
\subsubsection{High-Coverage Parallel Corpora}
Large-scale parallel corpora play an important role in developing machine translation systems, especially for low-resource languages. We have considered a dataset to be  high-coverage or large-scale if it contains more than 10 million sentence pairs, as this volume provides sufficient linguistic diversity and contextual richness for training robust translation models. These corpora serve as the backbone for both statistical and NMT systems, enabling improved generalization, domain adaptation, and cross-lingual transfer learning. 
BPCC Parallel Corpus ~\cite{gala2023indictrans2} stands out as the largest, containing 230 million sentence pairs across 22 Indic languages. Its extensive coverage makes it an invaluable resource for multilingual translation tasks, particularly for high-quality English-Indic translations. In comparison, the Samanantar Parallel Corpus ~\cite{ramesh2023samanantarlargestpubliclyavailable}, introduced in 2021, includes 46 million sentence pairs between English and 11 Indic languages, along with an additional 82 million sentence pairs between Indic languages. While smaller than BPCC, Samanantar is unique in its extensive Indic-Indic translation pairs, making it highly valuable for intra-Indic translation tasks.\\
Another notable dataset, CCAligned ~\cite{el-kishky-etal-2020-ccaligned}, consists of over 100 million document pairs across 137 languages, making it one of the most extensive cross-lingual resources. However, this dataset requires extensive filtering to improve data quality before being used for training MT models. Despite its noisiness, its unparalleled scale and diversity make it useful for large-scale pretraining and domain adaptation. The OPUS corpus ~\cite{Tiedemann2012OPUS}, widely recognized as a comprehensive multilingual dataset, aggregates multiple parallel corpora across various domains and languages. It covers over 100 million sentence pairs across more than 50 languages, including Hindi, Bengali, Marathi, Tamil, Telugu, and Malayalam, and overlaps significantly with the WAT 2018 Parallel Corpus ~\cite{zhang2020parallelcorpusfilteringpretrained}, which contains 10–20 million sentence pairs specifically for Hindi-English and Bengali-English translations. While OPUS provides domain diversity and structured data, it primarily consists of pre-existing datasets, making it less novel compared to BPCC and Samanantar. \\
Beyond these major corpora, other high-coverage Indic datasets contribute significantly to MT research. The Bhasha Parallel Corpus ~\cite{mujadia2025bhashaversetranslationecosystem} includes 44 million sentence pairs across seven Indic languages, supporting cross-lingual and domain adaptation studies. Additionally, the M2M-100 dataset ~\cite{fan2020beyond}, developed by Meta AI, contains 12 million sentence pairs spanning over 100 languages, including several Indic languages. This dataset played a key role in the development of the M2M-100 translation model, which enables direct translation between non-English languages.\\
Several other corpora derived from Wikipedia and Common Crawl have also contributed to large-scale MT training. The WikiMatrix Corpus ~\cite{schwenk-etal-2021-wikimatrix}, developed by Meta, consists of parallel sentences extracted from Wikipedia using LASER-based sentence alignment ~\cite{Artetxe_2019}. It offers a vast number of sentence pairs across numerous languages, it includes around 3.5 million Hindi-English sentence pairs and around 7 millions of other indic language pairs, making it a valuable resource for training MT models ~\cite{niehues-waibel-2011-using}. However, its reliance on Wikipedia content means that the domain of the text is primarily encyclopedic, limiting its applicability for more conversational or domain-specific translations. Similarly, Wikititles ~\cite{liu2017learningcharacterlevelcompositionalityvisual}, another Wikipedia-based corpus, extracts bilingual and multilingual article titles from Wikipedia, For Hindi-English, the dataset contains approximately 1.3 million parallel titles. Compared to WikiMatrix, Wikititles provides smaller, well-aligned phrase pairs rather than full sentences, making it particularly useful for training models that focus on short-form content, such as entity names, search queries, or phrase-based MT systems. Due to its structure, Wikititles is less prone to misalignment errors than WikiMatrix but lacks sentence-level parallelism. In contrast, CCMatrix ~\cite{schwenk-etal-2021-wikimatrix}, developed by Meta, is a much larger dataset mined from the CommonCrawl web corpus ~\cite{panchenko2018buildingwebscaledependencyparsedcorpus}. It contains billions of parallel sentences across multiple languages, surpassing both WikiMatrix and Wikititles in sheer volume, making it a powerful resource for large-scale NMT training. However, its primary drawback is the noisiness of web-mined content, which often includes misaligned or irrelevant text pairs that require extensive filtering. Among these datasets, WikiMatrix offers a balanced trade-off between scale and accuracy, providing a large yet relatively clean dataset for training MT models, whereas Wikititles ensures precise alignment quality but is limited in scope due to its focus on article titles rather than full sentences. CCMatrix, on the other hand, offers the most extensive collection of parallel sentences but requires aggressive filtering to ensure usability. While BPCC holds the advantage in terms of sheer size and multilingual support, Samanantar's inclusion of Indic-Indic translation pairs makes it particularly valuable for intra-Indic translation tasks. Researchers must carefully analyze these corpora to determine the most suitable dataset for their translation models, ensuring an optimal balance between high-quality curated translations and large-scale mined data. The choice of corpus ultimately depends on the specific needs of an MT system—whether prioritizing size, quality, or domain coverage.
\subsection{Low-Coverage Parallel Corpora}
While large-scale parallel corpora provide extensive training data for MT, many Indic languages remain low-resource, lacking sufficient parallel data for robust model development. For this study, we define low-resource parallel corpora as datasets containing fewer than 10 million sentence pairs. These datasets are crucial for developing MT models for underrepresented Indic languages, particularly those that lack substantial digital text resources. Despite their smaller size, these corpora serve as valuable benchmarks for fine-tuning, domain adaptation, and zero-shot learning approaches in machine translation.\\
The IIT Bombay Parallel Corpus ~\cite{kunchukuttan-etal-2018-iit} is one of the most widely used low-resource datasets, containing 1.5 million sentence pairs for English-Hindi translation. The dataset is derived from news and government documents, making it well-suited for formal text translation but less effective for conversational and domain-specific tasks. Due to its clean alignment and high-quality translations, it is often used for benchmarking and fine-tuning NMT models for Hindi-English translation. For Bangla-English translation, the \texttt{BUET\-English\-Bangla\-Corpus} ~\cite{buet2021bangla} offers 2.7 million sentence pairs, primarily sourced from news articles, books, and religious texts. While smaller than large-scale corpora like BPCC or OPUS, BUET ~\cite{buet2021bangla} is an important resource for Bangla machine translation, particularly for formal and literary domains. The dataset provides high-quality bilingual sentence alignments, making it a valuable resource for training MT models that need precise and domain-specific translations. \\
For historical and classical language translation, the Itihasa Parallel Corpus ~\cite{itihasa2020sanskrit} is one of the few available datasets, offering 93,000 sentence pairs for English-Sanskrit translation. Given Sanskrit’s morphologically rich structure and complex syntax, this dataset provides a rare opportunity for training translation models in ancient and scholarly texts. Due to its small size, MT models trained on Itihasa rely on data augmentation techniques, such as back-translation and transfer learning, to improve performance. A major initiative for low-resource machine translation across multiple Indic languages is the TICO-19 dataset ~\cite{anastasopoulos-etal-2020-tico}. TICO-19 provides parallel sentence pairs across multiple underrepresented Indic languages, including Maithili, Manipuri, and Sindhi. Unlike many general-purpose corpora, TICO-19 is specifically designed for medical and technical translations, making it highly valuable for domain-specific machine translation models. Given the importance of healthcare communication in multilingual settings, this dataset plays a critical role in enabling low-resource language translation for public health applications.\\
Another key dataset is NLLB (No Language Left Behind) ~\cite{nllb2022}, which provides small-scale training data for multiple Indic languages, including Kashmiri, Maithili, and Bhojpuri. The NLLB project is part of Meta AI’s initiative to support low-resource language translation, aiming to improve direct translation between non-English language pairs. While individual language pairs in NLLB have limited sentence pairs, the dataset’s wide coverage across many underrepresented Indic languages makes it highly useful for zero-shot and few-shot learning applications in MT.\\
In contrast to large-scale corpora such as BPCC and Samanantar, these low-resource parallel datasets serve as critical benchmarks for low-resource Indic languages, enabling research in domain adaptation, transfer learning, and cross-lingual generalization. Given the scarcity of annotated parallel corpora for many Indic languages, data augmentation, back-translation, and synthetic data generation play a crucial role in improving translation quality for underrepresented languages.
\subsection{Multimodal Corpora}
Multimodal corpora extend beyond traditional text-based datasets by incorporating multiple data types, such as text, speech, and visual information, into a unified dataset ~\cite{baltrusaitis2019multimodal}. These corpora are instrumental in building more comprehensive MT models capable of handling real-world scenarios involving multiple modalities ~\cite{li2020multimodal}. IndicMultiModal~\cite{kothapalli2021indicmultimodal}, for instance, provides text, speech, and image datasets aligned across multiple Indic languages, supporting research in multimodal translation, speech synthesis, and cross-lingual retrieval. Such corpora are particularly beneficial for applications in digital accessibility ~\cite{sun2021digitalaccessibility}, interactive AI assistants. 
\subsubsection{Speech-to-Text corpora}
Speech-to-text align spoken language with its textual translation, making them essential for speech translation and automatic speech recognition (ASR) systems ~\cite{jouvet2019asr}. These datasets are particularly valuable for creating voice-enabled translation models and developing ASR systems for Indic languages ~\cite{sitaram2020asr}.\\
One of the important corpora in this category is CVIT-IIITH Mann ki Baat Corpus ~\cite{Philip_2021}, mined from Indian Prime Minister Narendra Modi’s Mann ki Baat speeches. Since these speeches are carefully prepared and delivered in formal Hindi with occasional English phrases, the dataset is well-suited for studying political speech translation and handling Hindi-English code-switching. However, given its highly structured nature, it may not fully capture the variability of spontaneous speech, which is often a challenge for ASR models. Compared to other datasets, this corpus is more domain-specific, focusing on political communication. A more extensive alternative is the PMIndia corpus, which, while not explicitly speech data, consists of transcriptions of spoken news content into text~\cite{haddow2020pmindiacollectionparallel}. It extends beyond Mann ki Baat by providing spoken speech transcriptions with translations across multiple Indian languages. While both PMIndia and Mann ki Baat focus on government-related content, PMIndia includes a broader set of formal speeches, policy discussions, and governance-related material. This makes it valuable for multilingual speech translation systems, though, like Mann ki Baat, its primary limitation is that government discourse follows a standardized linguistic structure, lacking the variation seen in informal conversations or spontaneous speech.\\
Another dataset designed for a specific domain is the QED Corpus ~\cite{10.1162/tacl_a_00398}. It is a text data which is derived from the video transcripts. It focuses on educational video transcripts in English and Hindi. Which cover a wide range of topics, QED is optimized for academic discourse, including lecture-style content. This makes it particularly beneficial for ASR and translation models targeting online education platforms, academic lectures, and instructional content.  QED ensures high-quality transcriptions and translations tailored for educational use cases.
QED Corpus fills the gap in academic and instructional content, making it an essential resource for educational applications. The choice of corpus depends on the desired application—whether for structured government communication, spontaneous public discourse, or domain-specific speech processing.
\subsection{Text-to-Speech Corpora}
Text-to-Speech (TTS) technology plays an important role in enhancing accessibility and language inclusivity by converting textual information into natural-sounding speech. The development of high-quality TTS systems for Indic languages has gained momentum with the availability of large-scale corpora and open-source models.\\
Two notable contributions in this area are AI4Bharat Indic-TTS ~\cite{kumar2023buildingtexttospeechsystemsbillion} and BhasaAnuvaad ~\cite{jain2024bhasaanuvaad}, both of which provide extensive linguistic resources to improve speech synthesis and ASR systems. AI4Bharat Indic-TTS is an open-source TTS model that supports 13 Indic languages, including Assamese, Bengali, Gujarati, Hindi, Kannada, Malayalam, Marathi, Odia, Punjabi, Sanskrit, Tamil, Telugu, and Urdu. It is designed to produce high-quality synthetic speech with various speaker styles, making it suitable for applications such as language learning, assistive technologies, and media content creation. The model provides fine-grained control over speech parameters, including pitch, speed, and voice modulation, ensuring natural and expressive speech synthesis.
Complementing this, BhasaAnuvaad serves as a comprehensive dataset for speech translation, encompassing over 44,400 hours of speech and 17 million text segments across 13 Indic languages and English. By incorporating both mined and high-quality curated parallel speech data, BhasaAnuvaad is a valuable resource for developing ASR and TTS systems, as it enables sentence-to-audio alignment, facilitating the training of robust speech synthesis models. These initiatives significantly contribute to advancing TTS technology for Indic languages, fostering inclusivity and accessibility for a diverse user base.

\subsubsection{Image-to-Text Corpora}
Image-to-text corpora helps in advancing multimodal learning, enabling models to understand and generate text based on visual input ~\cite{DBLP:journals/cbm/GuoWSYCLZYXB24}. In the context of Indic languages, several high-quality datasets have been developed to support multimodal research, particularly in image captioning, visual question answering (VQA), and image-grounded translation ~\cite{zdemir2024EnhancingVQ}. Among the most significant contributions are the Hindi Visual Genome, Bengali Visual Genome, and Malayalam Visual Genome datasets. The Hindi Visual Genome dataset ~\cite{parida2019hindivisualgenomedataset} contains 31K multimodal pairs aligned in Hindi-English, providing a rich resource for training models in tasks such as image captioning and cross-lingual understanding. Similarly, the Bengali Visual Genome ~\cite{Sen2021BengaliVG} offers 29K multimodal pairs in Bengali-English, while the Malayalam Visual Genome includes 29K multimodal pairs in Malayalam-English. These datasets are designed to bridge the gap in low-resource Indic languages for multimodal AI applications. They are particularly useful in training ~\cite{Xue}  and evaluating MT models, cross-lingual retrieval systems, and vision-language models (VLMs) ~\cite{bordes2024introductionvisionlanguagemodeling} for Indic languages. These corpora contribute significantly to the development of multimodal AI for Indic languages, facilitating better captioning, improved VQA systems, and enhanced multilingual vision-language applications. By providing a strong benchmark for multimodal learning, they enable robust model training for real-world applications such as automated image description generation and visual assistive technologies in Indian languages.
\subsection{Code-Switched Corpora}
Code-switching, the practice of mixing two or more languages within a single conversation or sentence, is common in multilingual communities, including those using Indic languages ~\cite{garg2021codeswitching}. Code-switched corpora are crucial for developing translation models that can handle real-world conversations where users frequently switch between languages such as Hindi-English, Bengali-English, and Tamil-English ~\cite{bali2014codeswitching}. \\
GLUECoS ~\cite{khanuja2020gluecos} is a well known code-switched corpus, which contains Hindi-English and Bengali-English code-switched data and also datasets from social media platforms like Twitter and WhatsApp, where code-mixing is prevalent ~\cite{sitaram2019socialmedia}. Training MT systems on such corpora improves their ability to handle informal and conversational text ~\cite{winata2021codeswitching}.
The PHINC (Parallel Hinglish Social Media Code-Mixed Corpus) ~\cite{srivastava-singh-2020-phinc} consists of 13.7k Hinglish (Hindi-English) sentences, making it one of the most comprehensive resources for studying mixed-language usage in digital communication. It is particularly valuable for social media NLP tasks, where speakers often switch between Hindi and English within a single sentence.\\
Similarly, the IIIT-H en-hi-codemixed-corpus ~\cite{dhar-etal-2018-enabling} is a code-mixed dataset consisting of 6k English-Hindi sentences. Compared to PHINC, this corpus has a smaller dataset size but higher-quality annotation, ensuring accurate training data for models dealing with Hinglish content. Its focus on token-level annotations makes it especially useful for tasks such as word-level language identification and code-mixed text normalization. The CALCS 2021 Eng-Hinglish dataset ~\cite{appicharla-etal-2021-iitp} provides 10k parallel sentence pairs, focusing on formal and informal contexts of Hinglish usage. Compared to PHINC and IIIT-H, CALCS is particularly valuable for machine translation between English and Hinglish, helping models bridge the gap between standard English and code-mixed vernacular speech.

\section{Evaluation of Parallel Corpora}
\subsection{Evaluation Metrics}
Evaluating parallel corpora for Indic languages requires a combination of automatic evaluation metrics that compare machine-generated translations with human reference translations. Common metrics like BLEU ~\cite{papineni-etal-2002-bleu}, METEOR ~\cite{banerjee-lavie-2005-meteor}, COMET ~\cite{rei-etal-2020-comet}, and Translation Edit Rate (TER) ~\cite{snover-etal-2006-study} are frequently used, each offering different perspectives on translation quality. BLEU, which focuses on n-gram precision ~\cite{callison-burch-etal-2006-evaluating}, is a widely used metric but can struggle with Indic languages due to their lexical complexity and flexible word order. For instance, languages like Hindi and Tamil exhibit significant syntactic differences from English, which BLEU may fail to capture adequately. METEOR improves upon BLEU by incorporating recall, synonym matching, and stemming, which makes it more effective for languages with rich morphology and varied word forms, such as Hindi, Bengali, and Telugu ~\cite{li2024cultureparkboostingcrossculturalunderstanding}. However, like BLEU, METEOR still struggles with capturing semantic meaning and context, which are crucial for languages with high syntactic divergence from English.\\
To address these limitations, COMET, a newer metric, utilizes neural embeddings and contextualized models to evaluate translations based on semantic similarity and contextual understanding ~\cite{sun2024textualsimilaritykeymetric}. This makes COMET particularly valuable for Indic languages, where it is essential to capture contextual meaning and semantic equivalence rather than just surface-level n-gram matches ~\cite{rei-etal-2020-comet}. Despite its advantages, COMET requires pretrained models and significant computational resources, which may not always be feasible for resource-constrained settings ~\cite{larionov2024xcometlitebridginggapefficiency}. Additionally, TER measures the edit distance between the machine-generated translation and the reference ~\cite{stanchev-etal-2019-eed}. By counting the minimum number of insertions, deletions, or substitutions needed to transform one translation into the other, TER is particularly useful for identifying structural mismatches between languages, especially those with flexible sentence structures, such as Hindi and Tamil ~\cite{snover-etal-2006-study}. However, TER focuses on structural alignment rather than semantic accuracy, making it a complementary metric rather than a stand-alone tool.

\subsection{Human-Translated Evaluation Datasets}
High-quality, human-translated evaluation datasets are essential for benchmarking machine translation models, ensuring that they are assessed on accurate and reliable translations ~\cite{yan2024benchmarkinggpt4humantranslators}. Unlike automatically mined corpora, these datasets are manually curated by professional translators, making them gold-standard resources for evaluating translation quality across different language pairs. Human-translated corpora are particularly crucial for low-resource languages, where the availability of clean, parallel data is often limited ~\cite{haddow-etal-2022-survey}.\\
Several notable human-annotated evaluation datasets have been developed to facilitate rigorous benchmarking of multilingual machine translation systems. Among these, the FLORES-101 dataset ~\cite{guzman2019flores}, developed by Meta AI, is one of the most comprehensive evaluation resources. It provides human-translated test sets for 101 languages, including 14 Indic languages, making it a critical benchmark for assessing translation models in diverse linguistic settings. Following its success, Meta AI expanded the dataset to FLORES-200 ~\cite{guzman2019flores}, covering 200 languages, including 24 Indic languages. FLORES-200 represents one of the largest human-annotated evaluation datasets for multilingual translation, allowing researchers to systematically analyze the performance of models across a wide range of linguistic families. Both FLORES-101 and FLORES-200 use n-way parallel translation, meaning each sentence is consistently translated across all supported languages, enabling direct multilingual comparisons. These datasets are highly useful for benchmarking, but due to their relatively small sentence count, they are not suited for large-scale training.\\
Meta AI had also introduced No Language Left Behind (NLLB) ~\cite{nllbteam2022languageleftbehindscaling} a benchmarks for large-scale translation efforts,  NLLB-Seed ~\cite{nllb2022}, a small but valuable human-translated dataset specifically designed for evaluating very low-resource languages. This dataset focuses on five Indian languages—Kashmiri, Manipuri, Maithili, Bhojpuri, and Chhattisgarhi—where high-quality parallel data is scarce. While FLORES-200 provides extensive language coverage, it does not always include languages with very limited training data. NLLB-Seed ~\cite{nllb_seed_2022} fills this gap by prioritizing data quality and focusing on extremely low-resource languages, ensuring that translation models trained on scarce data sources can still be evaluated effectively. A complementary dataset, NLLB-MD (Multi-Domain) ~\cite{nllbteam2022languageleftbehindscaling}, extends this effort by providing human-annotated parallel translations across three key domains: news, unscripted informal speech, and health. Unlike FLORES datasets, which contain mostly general-purpose text, NLLB-MD allows for more fine-grained evaluation of translation models across different content types, addressing challenges such as domain adaptation and stylistic variation in machine translation. This makes it a valuable resource for improving translation models that operate in specific fields such as journalism, healthcare, or conversational AI. 

\subsection{Domain Adaptation and Bias}
The usability of parallel corpora for MT is contingent on their domain coverage and representativeness ~\cite{labaka-etal-2016-domain}. Most of the Indic corpora like samantar, PMIndia exhibit a strong bias toward formal and government-regulated domains, including legislative proceedings, religious scriptures, and legal texts ~\cite{khanuja2020gluecos}. While these datasets facilitate structured translation tasks, they lack the coverage necessary for informal and domain-specific language variations essential in social media, e-commerce, and medical translations. For example, Specialized corpora, such as TICO-19 ~\cite{anastasopoulos-etal-2020-tico}, have been curated to enhance healthcare-related translation, significantly improving domain-specific MT performance. Expanding parallel corpora to include informal, code-switched datasets from platforms like Twitter, WhatsApp, and online forums is crucial for improving translation quality in conversational and low-resource settings. Beyond domain generalization, data size and language representation remain pivotal. While large-scale corpora such as Bhasha ~\cite{jain2024bhasaanuvaad} provide substantial bilingual sentence pairs, their distribution skews heavily toward high-resource languages like Hindi and Bengali, leaving low-resource languages such as Santali, Konkani, and Maithili significantly underrepresented ~\cite{resnik-1999-mining}. Addressing this imbalance requires techniques such as cross-lingual transfer learning, wherein models pretrained on high-resource Indic languages are fine-tuned on their low-resource counterparts ~\cite{lample2019crosslingualtransfer}. Gender bias, particularly in languages like bengali with grammatical gender agreement, often results in incorrect translations of gender-neutral references ~\cite{stanovsky-etal-2019-evaluating}. Furthermore, political biases inherent in government-curated corpora such as PMIndia and Tico-19 may introduce ideological skew, influencing translation fidelity. Addressing such biases necessitates adversarial debiasing strategies, including counterfactual translation augmentation, reinforcement learning-based neutralization, and bias-aware adversarial training ~\cite{sun-etal-2019-mitigating}. Ensuring fair and inclusive translations mandates continuous evaluation and mitigation of systemic biases, particularly for Indic languages with diverse sociopolitical and linguistic landscapes.

\section{Future Directions}
The development of parallel corpora for Indic languages has advanced significantly, yet challenges related to data scarcity, domain diversity, and alignment quality persist. Future research must focus on expanding low-resource language coverage, improving domain adaptation, leveraging multimodal data, and enhancing automatic data generation techniques to build more robust MT and NLP systems for Indic languages.
\subsection{Expanding Coverage for Low-Resource and Dialectal Variants}
Many Indic languages, such as Santali, Bodo, Manipuri, and Konkani, remain underrepresented in existing parallel corpora. Most datasets focus on high-resource languages like Hindi, Bengali, and Tamil, creating an imbalance that hinders the development of MT models for low-resource languages ~\cite{lupascu2025largemultimodalmodelslowresource}. Future efforts should prioritize the collection of bilingual and multilingual parallel data from vernacular media, government archives, social media, and oral histories ~\cite{guzman2019flores}. Crowdsourcing initiatives and community-driven data curation can further help improve linguistic diversity and increase the representation of marginalized languages in NLP applications.
\subsection{Enhancing Domain Adaptation and Contextual Alignment}
Most existing Indic parallel corpora are domain-specific, with a strong bias toward news, religious texts, and government documents. This limits their applicability in scientific, medical, legal, and conversational domains ~\cite{hu2024gentranslatelargelanguagemodels}. To improve cross-domain generalization, future work should focus on constructing multi-domain parallel corpora and training domain-adaptive MT models ~\cite{dong2025advancesmultimodaladaptationgeneralization}. Furthermore, context-aware alignment techniques, such as document-level parallel corpora and sentence embedding-based alignment, can enhance semantic consistency and translation fluency across diverse textual genres.
\subsection{Leveraging Multimodal and Code-Switched Parallel Data}
Multimodal MT, which involves image-text and speech-text parallel corpora, is becoming increasingly relevant for Indic languages. Initial datasets like Hindi Visual Genome and Bengali Visual Genome demonstrate the potential of multimodal learning, but larger, more diverse datasets are needed to improve multimodal translation systems ~\cite{Sen2021BengaliVG}. Similarly, code-switching is prevalent in Hindi-English, Bengali-English, and Tamil-English interactions, yet parallel code-switched corpora remain scarce. Expanding multimodal and code-switched datasets will enhance MT models’ performance in real-world multilingual communication and improve their ability to handle informal language ~\cite{winata2021multimodal}.
\subsection{Automatic Data Generation}
Given the scarcity of human-annotated parallel corpora, automatic data generation techniques such as back-translation, synthetic data augmentation, and parallel data mining have gained prominence ~\cite{shu2024transcendinglanguageboundariesharnessing}. Back-translation, where monolingual target-language data is translated into the source language using pre-trained models, has been widely used to augment data for low-resource Indic languages ~\cite{sennrich2016backtranslation}. Similarly, parallel sentence mining techniques, such as LASER have been applied to extract sentence-aligned parallel data from large-scale web corpora. Additionally, zero-shot learning and self-supervised approaches can further help bootstrap translation models for languages with minimal parallel data. Future work should focus on refining these techniques to improve alignment accuracy and reduce noise in automatically generated corpora.

\section{Conclusion}
This paper has provided a comprehensive overview of parallel corpora for Indic languages, emphasizing their role in improving MT performance. While large-scale datasets like BPCC and Samanantar exist, many languages remain underrepresented, necessitating more diverse and high-quality resources. Challenges such as lexical diversity, script variations, and data scarcity require innovative approaches like crowd-sourcing, domain-specific text collection, and multimodal resources. Code-switching in digital communication also demands corpora that capture informal and mixed-language text. Automatic data generation techniques like back-translation and parallel sentence mining help augment corpora, but ensuring data quality is critical. Future research should address domain bias, improve evaluation metrics, and expand multimodal and low-resource language coverage. Collaborative efforts between researchers and linguistic communities will be essential in enhancing accessibility and translation accuracy for Indic languages.
\section{Limitations}
Despite providing a comprehensive review of parallel corpora for low-resource Indic languages in machine translation, several limitations must be acknowledged. The scope of the review is constrained by the availability of datasets, and while we have made an effort to cover a wide range of resources, many low-resource languages remain underrepresented. Additionally, the quality of the corpora varies significantly, with some datasets suffering from issues like inconsistent translations, noisy data, and domain-specific biases, which could limit their applicability in building robust machine translation systems. Moreover, this review focuses primarily on the datasets themselves and does not delve deeply into the models or evaluation metrics employed, which are crucial factors in the effectiveness of any MT system. Finally, access to some corpora may be restricted due to licensing issues, and in some cases, dataset metadata may not be fully available, limiting the depth of evaluation that can be performed. These limitations highlight the need for ongoing research and continuous updates in this evolving field.


\bibliography{main}
\nocite{Ando2005,andrew2007scalable,rasooli-tetrault-2015}
\appendix
\section{Appendix}
\label{sec:appendix}

\begin{table*}[ht!]
\centering
\renewcommand{\arraystretch}{1.1}
\resizebox{\textwidth}{!}{%
\begin{threeparttable}
\begin{tabular}{l|r|r|l|p{3.8cm}} 
\toprule
\textbf{Corpus Name} & \textbf{Corpus Type} & \textbf{Sentence Pairs} & \textbf{Languages} & \textbf{Link} \\
\midrule
\texttt{BPCC} & Text-to-Text & 230M & English-22 Indic languages & \href{https://ai4bharat.iitm.ac.in/datasets/bpcc}{BPCC} \\
\texttt{Samanantar} & Text-to-Text & 46M En-IL, 82M IL-IL & 11 Indic languages & \href{https://huggingface.co/datasets/ai4bharat/samanantar}{Samanantar} \\
\texttt{IIT Bombay} & Text-to-Text & 1.5M & English-Hindi & \href{https://www.cfilt.iitb.ac.in/iitb_parallel/}{IIT Bombay} \\
\texttt{CVIT-IIITH PIB} & Text-to-Text & N/A & Several Indic languages & \href{https://huggingface.co/datasets/jerin/pib}{CVIT-IIITH PIB} \\
\texttt{OPUS} & Text-to-Text & 100M+ & 50+ languages, several Indic & \href{http://opus.nlpl.eu/}{OPUS} \\
\texttt{WAT 2018} & Text-to-Text & 10-20M+ & Hindi-English, Bengali-English & \href{http://www.cs.umd.edu/~sagaw/projects/wat-2018.html}{WAT 2018} \\
\texttt{CCAligned} & Text-to-Text & 100M+ & 137 languages & \href{https://statmt.org/cc-aligned/}{CCAligned} \\
\texttt{Bhasha} & Text-to-Text & 44M+ rows & 7 Indic languages & \href{https://soket.ai/blogs/bhasha_wiki}{Bhasha} \\
\texttt{M2M-100} & Text-to-Text & 12M+ & 100+ languages & \href{https://github.com/facebookresearch/fairseq/tree/main/examples/m2m_100}{M2M-100} \\
\texttt{Itihasa} & Text-to-Text & 93K & English-Sanskrit & \href{https://github.com/rahular/itihasa}{Itihasa Corpus} \\
\texttt{BUET Eng-Bn Corpus} & Text-to-Text & 2.7M & English-Bangla & \href{https://github.com/csebuetnlp/banglanmt}{BUET Corpus} \\
\texttt{Sanskrit-Hindi-MT} & Text-to-Text & N/A & Sanskrit-English, Sanskrit-Hindi & \href{https://github.com/priyanshu2103/Sanskrit-Hindi-Machine-Translation}{Sanskrit-Hindi MT} \\
\texttt{Kangri Corpus} & Text-to-Text & 27,362 & Hindi-Kangri & \href{https://github.com/chauhanshweta/Kangri_corpus}{Kangri Corpus} \\
\texttt{MTEnglish2Odia} & Text-to-Text & 42K & English-Odia & \href{https://github.com/soumendrak/MTEnglish2Odia}{MTEnglish2Odia} \\
\texttt{IndoWordNet} & Text-to-Text & 6.3M & 18 Indic languages & \href{https://github.com/anoopkunchukuttan/indowordnet_parallel}{IndoWordNet Corpus} \\
\midrule
\texttt{NLLB Seed} & Text-to-Text & N/A & Kashmiri, Maithili, Bhojpuri & \href{https://github.com/facebookresearch/flores/tree/main/nllb_seed}{NLLB Seed} \\
\texttt{PHINC} & Text-to-Text & 13,738 & Hindi-English Code-mixed & \href{https://www.aclweb.org/anthology/2021.findings-acl.253/}{PHINC} \\
\texttt{NLLB MD} & Text-to-Text & 9000+ & Bhojpuri & \href{https://github.com/facebookresearch/flores/blob/main/nllb_md/README.md}{NLLB-MD} \\
\texttt{PMIndia} & Text-to-Text & N/A & Hindi-English & \href{https://cvit.iiit.ac.in/}{PMIndia} \\
\texttt{QED} & Text-to-Text & 43K & English-Hindi & \href{https://alt.qcri.org/resources/qedcorpus/}{QED Corpus} \\
\texttt{CoPara} & Text-to-Text & 2.5K passage pairs & 4 Dravidian languages & \href{https://github.com/ENikhil/CoPara}{CoPara} \\
\texttt{Uka Tarsadia} & Text-to-Text & 65K & English-Gujarati & \href{https://github.com/shahparth123/eng_guj_parallel_corpus}{Uka Tarsadia} \\
\texttt{TICO 19} & Text-to-Text & N/A & Multiple indic languages & \href{https://tico-19.github.io/}{TICO 19} \\
\midrule
\texttt{BhasaAnuvaad} & Speech<->Text & 44,400+ hrs & 13 Indic languages & \href{https://github.com/AI4Bharat/BhasaAnuvaad}{BhasaAnuvaad} \\
\texttt{Mann ki Baat} & Speech<->Text & N/A & Hindi & \href{https://www.kaggle.com/datasets/taruntiwarihp/pm-india-mann-ki-baat}{Mann ki Baat} \\
\texttt{IndicTTS} & Speech<->Text & 100+ hrs/language & 7 Indic languages & \href{https://github.com/AI4Bharat/Indic-TTS}{IndicTTS} \\
\midrule
\texttt{GLUECoS} & Code-Switched & 8K-22K & Hindi-English Code-mixed & \href{https://microsoft.github.io/GLUECoS/}{GLUECoS} \\
\texttt{PHINC} & Code-Switched & 13,738 & Hindi-English Code-mixed & \href{https://aclanthology.org/2020.wnut-1.7/}{PHINC} \\
\texttt{IIIT-H en-hi-codemixed} & Code-Switched & 6K & English-Hindi & N/A \\
\texttt{CALCS 2021} & Code-Switched & 10K & English-Hinglish & \href{https://ritual.uh.edu/lince/datasets}{CALCS 2021} \\
\midrule
\texttt{Hi Visual Genome} & Multimodal & ~31K & Hindi-English & \href{https://arxiv.org/abs/1907.08948}{Hindi Visual Genome} \\
\texttt{Bn Visual Genome} & Multimodal & 29K & Bengali-English & \href{https://link.springer.com/chapter/10.1007/978-981-16-6624-7_7}{Bengali Visual Genome} \\
\texttt{Ml Visual Genome} & Multimodal & 29K & Malayalam-English & \href{https://lindat.mff.cuni.cz/repository/xmlui/handle/11234/1-3533}{Malayalam Visual Genome} \\
\bottomrule
\end{tabular}
\end{threeparttable}
}
\vspace{1mm}
\caption{
Comprehensive overview of major parallel corpora available for Indic languages, spanning various modalities including text-to-text, speech-to-text, code-switched, and multimodal datasets. The table highlights the corpus type, number of sentence pairs or duration (where applicable), supported language pairs (with a focus on English-Indic and intra-Indic combinations), and links to official sources for access. These resources play a vital role in enabling research in machine translation, multilingual NLP, and low-resource language processing across the Indic language spectrum.
}
\label{Tab:parallel-corpora}
\end{table*}

\begin{table*}[ht!]
  \label{Tbl:indic-scripts}
  \centering
  \resizebox{2.2\columnwidth}{!}{%
  \begin{threeparttable}
  \begin{tabular}{l|l|l|p{3.8cm}}
    \toprule
    \textbf{Script} & \textbf{Languages} & \textbf{Region} & \textbf{Script Family} \\
    \midrule
    Devanagari & Hindi, Marathi, Sanskrit, Nepali, Konkani, Maithili, Bhojpuri, Sindhi & North, Central India, Nepal & Brahmic \\
    \midrule
    Bengali & Bengali, Assamese, Sylheti, Bodo & Eastern India, Bangladesh & Brahmic \\
    \midrule
    Sharada & Kashmiri (historical script) & Kashmir (historical) & Brahmic \\
    \midrule
    Gurmukhi & Punjabi & Punjab (India and Pakistan) & Brahmic \\
    \midrule
    Gujarati & Gujarati & Gujarat, Daman and Diu & Brahmic \\
    \midrule
    Odia & Odia & Odisha & Brahmic \\
    \midrule
    Grantha & Tamil (Sanskrit texts), Kannada & Tamil Nadu, Karnataka (historical) & Brahmic \\
    \midrule
    Tamil & Tamil & Tamil Nadu, Sri Lanka, Singapore & Brahmic \\
    \midrule
    Telugu & Telugu & Andhra Pradesh, Telangana & Brahmic \\
    \midrule
    Kannada & Kannada & Karnataka & Brahmic \\
    \midrule
    Malayalam & Malayalam & Kerala & Brahmic \\
    \midrule
    Urdu (Arabic script) & Urdu, Kashmiri, Dakhini & North India, Pakistan, Kashmir & Arabic \\
    \midrule
    Arabic & Arabic, Sindhi & North India, Pakistan, Jammu & Kashmir \\
    \midrule
    Tibetan & Tibetan (spoken in Ladakh, Sikkim) & Ladakh, Sikkim, Tibet & Tibetic \\
    \midrule
    Meitei Mayek & Manipuri & Manipur & Brahmic \\
    \midrule
    Brahmi & Ancient Indian texts, Prakrits, early Sanskrit & Pan-Indian (historical) & Brahmic \\
    \midrule
    Sinhala & Sinhala & Sri Lanka (but used by Tamil diaspora in India) & Brahmic \\
    \midrule
    Lepcha & Lepcha & Sikkim, Darjeeling & Brahmic \\
    \midrule
    Limbu & Limbu & Sikkim, Darjeeling, eastern Nepal & Brahmic \\
    \midrule
    Tirhuta & Maithili & Bihar, Nepal & Brahmic \\
    \midrule
    Kaithi & Hindi (historical script) & Bihar, Uttar Pradesh & Brahmic \\
    \midrule
    Sylheti Nagari & Sylheti & Bangladesh, India (Assam) & Brahmic \\
    \midrule
    Chakma & Chakma & Chittagong Hill Tracts (Bangladesh) & Brahmic \\
    \midrule
    Burmese & Burmese & Myanmar, parts of India (Mizoram) & Burmese \\
    \midrule
    Thai & Thai & Thailand (historical influence in India) & Thai \\
    \midrule
    Khmer & Khmer (used historically in Southeast India) & Cambodia, some historical presence in India & Khmer \\
    \bottomrule
  \end{tabular}
  \end{threeparttable}
  }
 \caption{
Comprehensive overview of the diverse scripts used across Indic languages, categorized by associated languages, geographic regions, and their respective script families. This table, referenced in Section~\ref{subsec:challenges}, highlights both modern and historical scripts, including Brahmic-derived scripts (e.g., Devanagari, Tamil, Bengali), Perso-Arabic adaptations (e.g., Urdu, Kashmiri), and lesser-known indigenous scripts (e.g., Meitei Mayek, Lepcha, Chakma). The representation illustrates the linguistic diversity and orthographic complexity of the Indian subcontinent—factors that critically affect text normalization, OCR development, and
}
\end{table*}

\begin{figure*}[h!]
\begin{minipage}[b]{1.0\linewidth}
  \centering
  \centerline{\includegraphics[width=15.0cm]{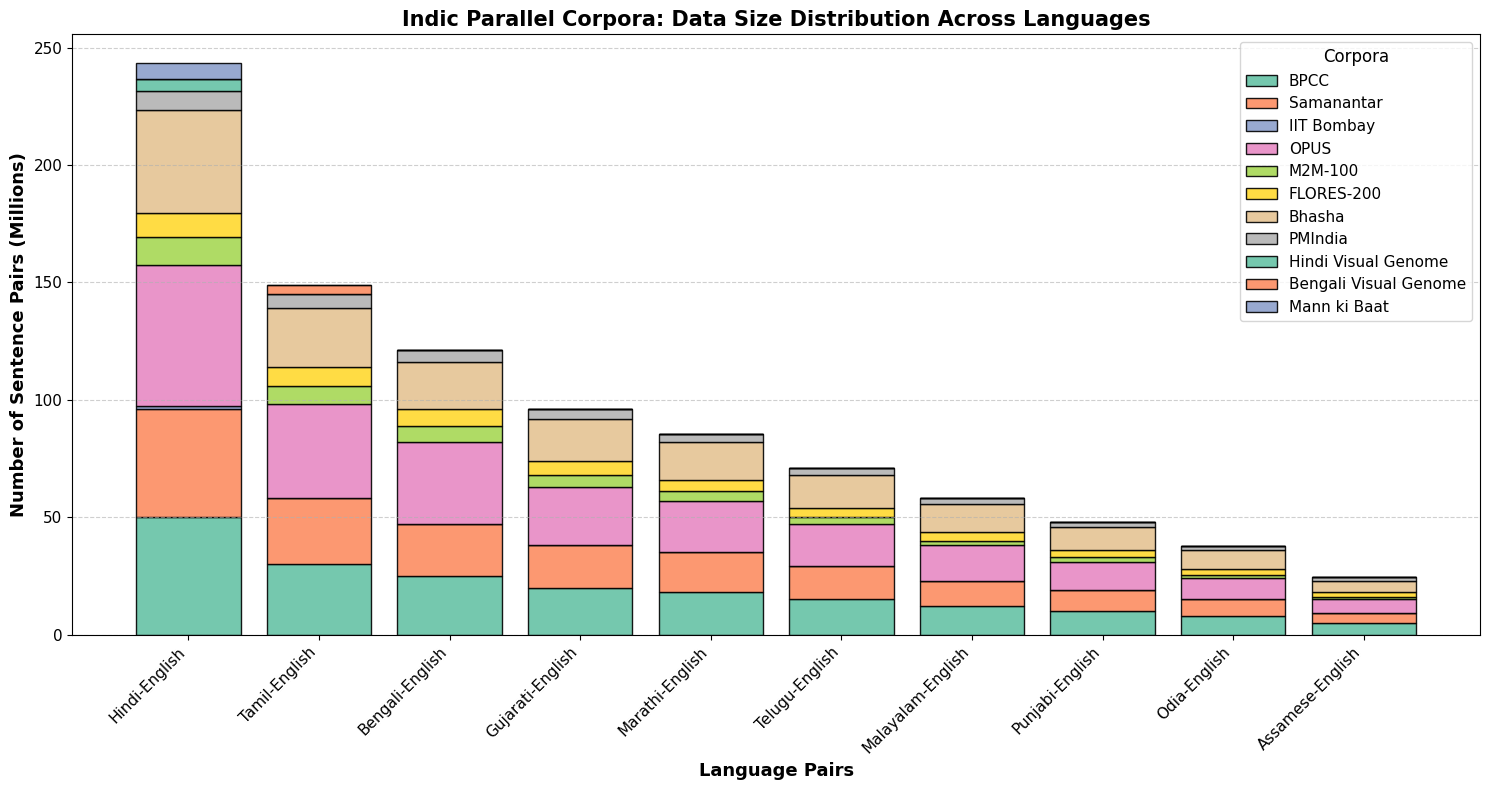}}
  \caption{\label{fig_data} Indic Parallel Corpora: Data Size Distribution Across Languages. A stacked bar chart showing the number of sentence pairs (in millions) for various Indic-English language pairs across major corpora. Languages like Hindi, Tamil, and Bengali are well-resourced, while others such as Assamese and Odia have significantly less data. This highlights the data imbalance in Indic NLP and the need for better resource coverage.}
\end{minipage}
\end{figure*}

\end{document}